\documentclass{article}
\usepackage{amsmath,epsfig}
\usepackage[preprint]{spconfa4}
\usepackage{xcolor}
\usepackage{cite}

\let\OLDthebibliography\thebibliography
\renewcommand\thebibliography[1]{
  \OLDthebibliography{#1}
  \setlength{\parskip}{0pt}
  \setlength{\itemsep}{0pt plus 0.3ex}
}

\usepackage[plain,noend,ruled]{algorithm2e}
\usepackage{booktabs}
\usepackage{amsmath}
\usepackage{amssymb}
\usepackage{mathrsfs}
\usepackage{amsthm}
\usepackage{enumerate}

\newcommand{\BA}{\mathbf{A}}    \newcommand{\BB}{\mathbf{B}}

\theoremstyle{plain}
\newtheorem{assumption}{Assumption}

\begin{document}\sloppy


\title{IMPROVING DEEP IMAGE MATTING VIA LOCAL SMOOTHNESS ASSUMPTION}
%
\name{Rui Wang$^1$, Jun Xie$^{2\ast}$\thanks{$^{\ast}$Corresponding author: Jun Xie  (xiejun@lenovo.com)}, Jiacheng Han$^3$ and Dezhen Qi$^4$}
\address{
$^1$Center for Applied Statistics and School of Statistics,
\\
Renmin University of China,
Beijing 100872, China
\\
$^2$PCIE Lab, Lenovo Research, Beijing, China
\\
$^3$School of Optics and Photonics, Beijing Institute of Technology, Beijing, China
\\
$^4$Department of Physics, and Fujian Provincial Key Laboratory for Soft Functional Materials Research,
\\
Xiamen University,
Xiamen 361005, China
}

\maketitle

\begin{abstract}
    Natural image matting is a fundamental and challenging computer vision task. Conventionally, the problem is formulated as an underconstrained problem. Since the problem is ill-posed, further assumptions on the data distribution are required to make the problem well-posed. For classical matting methods, a commonly adopted assumption is the local smoothness assumption on foreground and background colors. However, the use of such assumptions was not systematically considered for deep learning based matting methods. In this work, we consider two local smoothness assumptions which can help improving deep image matting models. Based on the local smoothness assumptions, we propose three techniques, i.e., training set refinement, color augmentation and backpropagating refinement, which can improve the performance of the deep image matting model significantly. We conduct experiments to examine the effectiveness of the proposed algorithm. The experimental results show that the proposed  method has favorable performance compared with existing matting methods.
\end{abstract}

\section{Introduction}

Natural image matting is an important task in image editing.
In this task,
the natural image $\mathcal I$ is assumed to be a convex combination of the
foreground $\mathcal F \in \mathbb R^{h \times w \times 3}$ and 
the background $\mathcal B \in \mathbb R^{h \times w \times 3}$
weighted by $\alpha \in [0, 1]^{h \times w}$, the opacity of the foreground.
Formally,
the color at $i \in \{1, \ldots, h\} \times \{1, \ldots, w\}$ satisfies the compositing equation
{\small
\begin{equation}
\label{eq:fundamental}
    \mathcal I_i^c= \alpha_i \mathcal F_i^c + (1-\alpha_i) \mathcal B_i^c,
    \quad c= 1,2, 3.
\end{equation}
}%
Given an input image $\mathcal I$,
the goal is to estimate $\alpha$.
Note that there are $3$ equations with $7$ unknowns at each pixel.
This results in the nonidentifiability of $\mathcal F$, $\mathcal B$ and $\alpha$.
Specifically, suppose that the triplet $(\mathcal F, \mathcal B, \alpha)$ satisfies \eqref{eq:fundamental}.
Then for any $\mathcal Q \in \mathbb R^{h \times w}$ satisfying $0 \leq \mathcal Q_i \leq \frac{1}{\alpha_i}$,
the triplet $((\mathcal B_i^c + (\mathcal F_i^c - \mathcal B_i^c) /\mathcal Q_i), \mathcal B, (\mathcal Q_i \alpha_i) )$ also satisfies \eqref{eq:fundamental}.
Thus, without further assumption, the natural image matting problem is ill-posed.

To regulate the problem,
natural image matting is often aided by human.
In a typical human-aided process,
the user provides a trimap $\mathcal T \in \{0, 0.5, 1\}^{h \times w}$ indicating
the purely foreground region $\{i: T_i = 1\}$, 
the purely background region $\{i: T_i = 0\}$,
and the unknown region $\{i: T_i = 0.5\}$.
It is guaranteed that if $\mathcal T_i = 1$ then $\alpha_i = 1$ and if $\mathcal T_i = 0$ then $\alpha_i = 0$.
With a trimap $\mathcal T$, one only needs to estimate $\alpha$ in the unknown region.
However, the compositing equation \eqref{eq:fundamental} is still ill-posed in the unknown region.
In principle, 
further assumptions are still required to make the problem well-posed.



For traditional image matting methods,
local smoothness assumptions on $\mathcal F$, $\mathcal B$ or $\alpha$ are commonly used.
In a seminal work,
Levin \emph{et al.} \cite{Levin2008AClosed-Form} proposed the color line model based on the assumption that $\mathcal F$ and $\mathcal B$ are approximately constant locally.
There are also methods which utilize smoothness assumptions on $\alpha$; 
see \cite{Price2010Simultaneous} and the references therein.
For traditional methods,
a main goal of
imposing local smoothness assumptions
is to formulate tractable models.
Consequently, their assumptions may be overly strong.

Thanks to the large training data released by Xu \emph{et al.} \cite{Xu2017DeepImageMatting}
and the advent of convolutional neural networks,
deep learning based matting methods have
achieved significantly better performances than traditional methods \cite{Xu2017DeepImageMatting,Li2020NaturalImageMatting,Tang2019Learning-Based,Forte2020FBA,Yu2021High-ResolutionDeepImageMatting,Sun2021SemanticImageMatting}.
For deep learning methods,
one does not need to formulate an explicit model of colors.
Rather, the deep neural networks may learn the implicit local smoothness assumptions during training.
Intuitively, deep learning methods may be improved if they are explicitly guided by local smoothness assumptions.
Unfortunately, to our best knowledge, no existing deep learning method utilizes the local smoothness assumptions explicitly.


\begin{figure*}[t]
  \centering
  \includegraphics[width = 1.95\columnwidth]{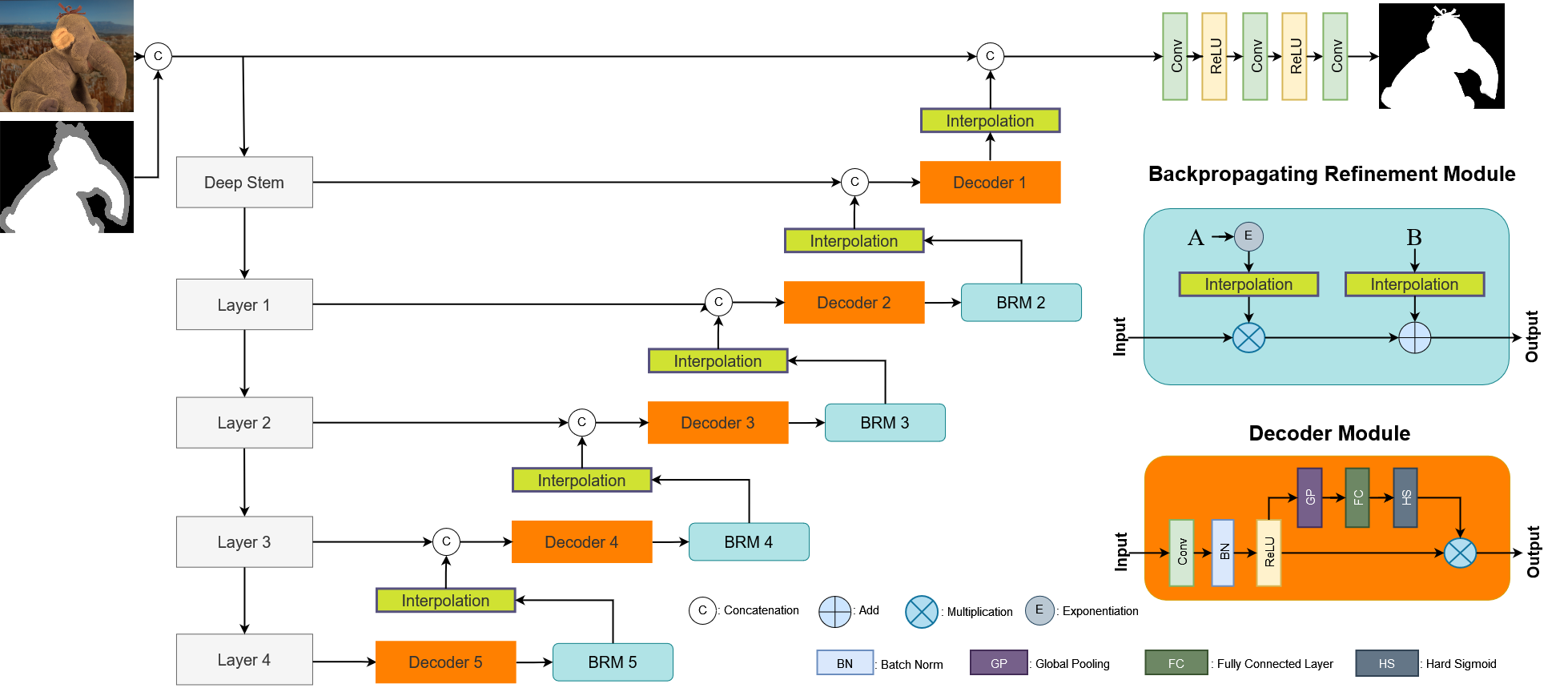}
  \caption{Proposed network architecture.}
  \label{fig:1}
\end{figure*}
In this paper, we explore the use of local smoothness assumption 
in deep image matting.
We consider two seemingly trivial assumptions that should be satisfied by natural images.
Surprisingly, these simple assumptions can be violated by modern deep image matting methods.
We propose three techniques to guide deep image matting methods based on the local smoothness assumpitons.
The proposed three techniques work in the different stages of deep learning methods.
We observe that the foreground images in the Adobe Image Matting (AIM) dataset \cite{Xu2017DeepImageMatting} violate our first assumption.
Hence our first technique is a training set refinement method which improves the AIM training set.
The second technique is a color augmentation method which functions in the training phase.
This technique can further improve the local smoothness property of the training data.
The third technique is a backpropagating refinement method which functions in the inference phase.
It enforces the output of the deep network to satisfy our second assumption.
We conduct experiments to evaluate the effectiveness of the proposed techniques.
In summary, the contributions of the present paper are as follows:
\begin{itemize}
    \item
        This work is the first one to systematically explore the local smoothness assumptions in deep image mattting.
    \item
        We propose three novel techniques to improve the performance of deep image matting models.
        In particular, we present the first deep image matting method which uses backpropagating refinement during inference.
    \item
        Based on the proposed techinques,
        we present a simple deep image matting model which achieves the state-of-the-art performance on the testing set of \cite{Xu2017DeepImageMatting} among all methods trained solely on AIM dataset.
\end{itemize}




\section{Related works}

A class of classical matting methods are based on pixel sampling.
These methods sample colors from the known foreground region and known background regions, and use them to estimate $\alpha_i$ in the unknown region.
Matting methods based on pixel sampling include
\cite{Chuang2001ABayesianApproach,Wang2007OptimizedColorSampling,Gastal2010SharedSampling,He2011AGlobalSampling,Shahrian2013ImprovingImageMatting,Feng2016AClusterSampling}.

Another traditional approach to the natural image matting problem is based on affinities between pixels.
These methods make use of pixel similarities to propagate the alpha values from the region of $\{i: \mathcal T_i = 0\} \cup \{i: \mathcal T_i = 1\}$ to the region of $\{i: \mathcal T_i = 0.5\}$.
Poisson matting mehtod \cite{Sun2004PoissonMatting} impose a model of the gradient of $\mathcal I$ and $\alpha$ and use Poisson equations to characterize the afinities between pixels.
Levin \emph{et al.} \cite{Levin2008AClosed-Form} proposed a color line model based on local smoothness assumptions and obtain a closed-form matting method.
Chen \emph{et al.} \cite{Chen2013KNNMatting} proposed a KNN matting method which incorporates nonlocal information.
Aksoy \emph{et al.} \cite{Aksoy2017DesigningEffective} conceptualized the affinities as information flows and designed several information flows to control the way of propagation.

The deep learning method was introduced to the matting problem by
Cho \emph{et al.} \cite{Cho2016NaturalImageMatting}. 
Xu \emph{et al.} \cite{Xu2017DeepImageMatting} released a large scale matting dataset which greatly facilitated the subsequent deep learning methods for natural image matting.
Lutz \emph{et al.} \cite{Lutz2018AlphaGAN} explored adversarial training in deep image matting.
Lu \emph{et al.} \cite{Lu2019IndicesMatter} designed an upsampling module for matting.
Tang \emph{et al.} \cite{Tang2019Learning-Based} proposed a deep learning based sampling method.
Cai \emph{et al.} \cite{Cai2019Disentangled} disentangled the matting problem to  trimap adaptation and alpha estimation.
Li and Lu \cite{Li2020NaturalImageMatting} proposed a nonlocal module to capture long-range contexts.
Forte and Piti\'e \cite{Forte2020FBA} proposed a network which can simultaneously estimate $\mathcal F$, $\mathcal B$ and $\alpha$.
Yu \emph{et al.} \cite{Yu2021High-ResolutionDeepImageMatting} proposed a deep learning method to deal with images with extremely large size.
Sun \emph{et al.} \cite{Sun2021SemanticImageMatting} proposed a network which incorporates semantic classification.

Besides the deep learning methods mentioned above, there are also some deep learning methods which do not need an input trimap; see, e.g., \cite{Sengupta2020BackgroundMatting,Qiao2020Attention-GuidedHierarchical}.

\section{A Baseline Deep Image Matting Model}
In this section, we introduce a simple deep neural network for natural image matting which serves as the baseline model.

\textbf{Network architecture.}
Our network, as illustrated in Fig. \ref{fig:1},
has a standard encoder-decoder architecture.
The encoder is ResNet-50-D \cite{He2019BagOfTricks}, a variant of ResNet-50 \cite{He2016DeepResidual}.
The input of the encoder is the concatenation of $\mathcal I$ and $\mathcal T$, which has $4$ channels.
The input is $32\times$ downsampled in the encoder.
In the baseline model, the backpropagating refinement module is simply the identity operator.
In the decoder,
to fuse the global information of the feature map,
we adopt the blocks in VoVNet \cite{Lee2019AnEnergy} which are similar to the SE block \cite{Hu2020SqueezeAndExcitation}.
Compared with the SE block, 
our module has only one convolution and uses the hard sigmoid which is faster than sigmoid.


\textbf{Loss function.}
Xu \emph{et al.} \cite{Xu2017DeepImageMatting} considered the loss function
{\small
    $$
        L(\hat \alpha, \alpha) =
        \frac{1}{
            \mathrm{Card}
            (\{i: \mathcal T_i = 0.5\})
        }
        \sum_{\{i: \mathcal T_i = 0.5\}}
        \sqrt{(\hat \alpha_i - \alpha_i)^2 + \epsilon^2 }
        .
        $$
}%
To measure the difference of $\hat \alpha$ and $\alpha$ at different scales, we apply the above loss function at various scales of $(\hat \alpha, \alpha)$ and use their weighted sum as the final loss function.
Specifically, our loss function is
        $\sum_{\ell = 0}^4
        2^{-\ell} 
        L( \mathscr P^\ell(\hat \alpha), \mathscr P^\ell(\alpha) )
        $,
where $\mathscr P^\ell$ denotes the average pooling operator with kernel size $2^\ell \times 2^\ell$ and stride $2^\ell$.


\section{Utilizing Local Smoothness Assumptions}






For a position $i$, let $\partial \{i\}$ denote the set of $4$ surrounding positions of $i$.
For a region $R$, let $\partial R$ denote the boundary of $R$.
Formally, $i \in \partial R$ if and only if $(\{i\} \cup \partial \{i\}) \cap R \neq \emptyset$ and $(\{i\} \cup \partial \{i\}) \cap R^\complement \neq \emptyset$.
We consider the following two modest local smoothness assumptions.
\begin{assumption}\label{assumption1}
        $\mathcal F$ is locally constant at the positions
        {\small
            \begin{align*}
        \{i : \alpha_i = 0 \} \cup \partial \{i : \alpha_i = 0 \}.
    \end{align*}
        }%
\end{assumption}
\begin{assumption}\label{assumption2}
        $\alpha$ is locally constant at the positions
        {\small
            \begin{align*}
         \partial \{i: \mathcal T_i =1\} \cup \partial\{i: \mathcal T_i =0\}.
            \end{align*}
        }%
\end{assumption}
Assumptions \ref{assumption1} and \ref{assumption2} impose conditions on $\mathcal F$ and $\alpha$, respectively.
These assumptions are fairly weak.
We can expect them to hold for general natural images.
We do not impose any condition on $\mathcal B$.
In fact,
the background $\mathcal B$ is often an unconstrained natural image in practice.
%

To justify Assumption \ref{assumption1}, 
we note that according to the 
compositing equation \eqref{eq:fundamental},
the value of $\mathcal F$ in the region $\{i:\alpha_i = 0\}$ does not affect the color of $\mathcal I$, and hence can be arbitrarily defined.
However, the foreground colors in the region $\{i:\alpha_i = 0\}$ do have impact on the training process of deep image matting model.
In fact, some commonly adopted data augmentation techniques, e.g., image resizing and image rotation, rely on interpolation techniques.
When interpolation techniques are applied to $\mathcal F$,
the nonsmooth colors in the region $\{i:\alpha_i = 0\} \cup \partial \{i : \alpha_i = 0 \}$ may contaminate the foreground colors in the region $\{i:\alpha_i > 0\}$.
This phenomenon was observed by \cite{Forte2020FBA}.
The problem of color contamination is eased by Assumption \ref{assumption1}.
In fact, under Assumption \ref{assumption1},
for a position $i$ in $\{i:\alpha_i > 0\}$,
the sampling points of the interpolation method has similar foreground colors as $i$ and hence will not contaminate the color of $i$.

To justify Assumption \ref{assumption2},
we note that the trimap provided by users are often coarse.
Since the trimap must be precise, the users may tend to be conservative and do not label pixels near the region $\{ i: 0< \alpha_i < 1 \}$ to be foreground or background.
As a result, the pixels near the region $\{i: \mathcal T_i =1\}$ are very likely to have opacity exactly $1$, and the pixels near the region $\{i: \mathcal T_i = 0\}$ are very likely to have opacity exactly $0$.
Further, 
for the open matting datasets
of \cite{Rhemann2009APerceptually} and \cite{Xu2017DeepImageMatting},
the trimaps are based on dilations of the region
$\{ i: 0< \alpha_i < 1 \}$.
For these datasets, Assumption \ref{assumption2} holds strictly.
In summary, Assumption \ref{assumption2} is a weak and  reasonable assumption.

%
Below we propose three techniques based on Assumptions \ref{assumption1} and \ref{assumption2} to improve deep image matting.

\subsection{Refining Training Set}
Some commonly used data augmentation techniques in deep image matting rely on interpolation techniques.
As observed by Forte and Piti\'e \cite{Forte2020FBA},
when applying interpolation techniques to $\mathcal F$,
the nonsmooth colors in the region $\{i:\alpha_i = 0\} \cup \partial \{i : \alpha_i = 0 \}$ may contaminate the foreground colors in the region $\{i:\alpha_i > 0\}$.
That is, the violation of Assumption \ref{assumption1} results in color contamination.
To ease this problem,
Forte and  Piti\'e \cite{Forte2020FBA} used the closed-form foreground estimation method in \cite{Levin2008AClosed-Form} to re-estimate the foregrounds in AIM dataset.
However, The method of \cite{Levin2008AClosed-Form} was not designed to meet Assumption \ref{assumption1}.
We shall see that their method has suboptimal performance compared with the proposed re-estimation method.

\begin{figure*}[t]
  \centering
  \includegraphics[width = 1.9\columnwidth]{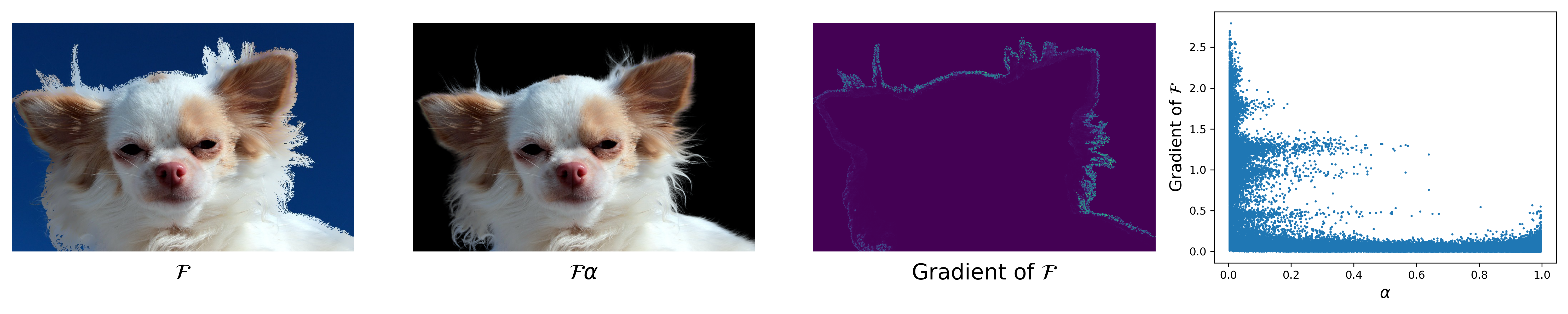}
  \\
  \includegraphics[width = 1.9\columnwidth]{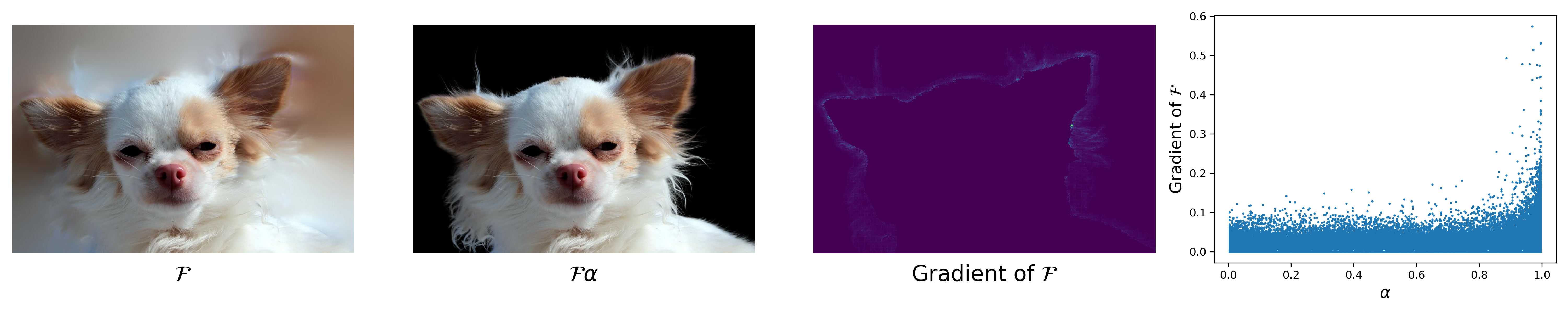}
  \caption{Illustration of the effect of re-estimation of $\mathcal F$. 
      Figures in the first row are for the original $\mathcal F$.
      Figures in the second row are for the re-estimated $\mathcal F$.
      The first column is the image $\mathcal F$.
      The second column is the image $\mathcal F \alpha$.
      The third column is the gradient of $\mathcal F$.
      The Fourth column is the plot of the gradient of $\mathcal F$ versus $\alpha$ within the region $\{i: 0 < \alpha_i < 1\}$.
  }
  \label{fig:four_and_four}
\end{figure*}






In the training set of AIM, $\mathcal F$ does not meet Assumption \ref{assumption1}.
Now we present a new method to re-estimate $\mathcal F$.
Formally, the re-estimation problem is as follow:
given an initial foreground $\mathcal F$ and $\alpha$, the goal is to output a re-estimated $\hat {\mathcal F}$ such that 
$\hat {\mathcal F}$
has a similar behavior as
${\mathcal F}$ when used to compose an image
and
that 
$\hat {\mathcal F}$ satisfies Assumption \ref{assumption1}. 
To meet these two requirements, we consider the following cost function:
{\small
    \begin{align*}
    C(\hat {\mathcal F}, \mathcal F)
    =&
    \frac{1}{2}
    \sum_{i \in \{1, \ldots, h\} \times \{1, \ldots, w\}}
    \sum_{c=1}^3
    \Big\{
        \alpha_i^2 ( \hat {\mathcal F}_i^c - {\mathcal F}_i^c )^2
        \\
     &
     \quad\quad\quad
     \quad\quad\quad
    +
    \kappa
    (1 - \alpha_i)^2
        \sum\limits_{j \in \partial \{i\} } (\hat {\mathcal F}_i^c - \hat {\mathcal F}_j^c)^2
    \Big\}
    ,
\end{align*}
}%
where $\kappa > 0$ is a hyperparameter.
The cost
$
        \alpha_i^2 ( \hat {\mathcal F}_i^c - {\mathcal F}_i^c )^2
$
ensures $\hat {\mathcal F}_i^c$ is close to ${\mathcal F}_i^c $ when $\alpha_i$ is large.
On the other hand, the cost
$
    (1 - \alpha_i)^2
        \sum_{j \in \partial \{i\} } (\hat {\mathcal F}_i^c - \hat {\mathcal F}_j^c)^2
$
ensures that $\hat{\mathcal F}_i^c$ is locally smooth when $\alpha_i $ is small, which conforms Assumption \ref{assumption1}. 
The derivative of $C( \hat{\mathcal F}, \mathcal F)$ is
{\small
\begin{align*}
    \frac{\partial C (\hat {\mathcal F}, \mathcal F) }{\partial \hat {\mathcal F}_i^c} = &
    \left(
        \alpha_i^2 
    +
    4\kappa (1-\alpha_i)^2
    +
    \kappa \sum\limits_{j\in \partial \{i\}} (1-\alpha_j)^2 
\right)
\hat {\mathcal F}_i^c
    \\
    &
    -\alpha_i^2 \mathcal {\mathcal F}_i^c
    -
     \kappa {\sum\limits_{j \in \partial\{i\} }
         (
(1-\alpha_i)^2
+(1-\alpha_j)^2
)
     \hat {\mathcal F}_j^c }
     .
\end{align*}
}%
Setting
the derivative $\partial C (\hat{\mathcal F}, \mathcal F) / \partial \hat {\mathcal F}_i^c $ to zero yields
{\small
\begin{align*}
    \hat {\mathcal F}_i^c = 
    \frac{
        \alpha_i
        \mathcal {\tilde I}_i^c
        +
        \kappa
         \sum_{j \in \partial \{i\}} ((1-\alpha_i)^2 +(1 -\alpha_j)^2 ) \hat {\mathcal F}_j^c
    }{
        \alpha_i^2 + 
        4 \kappa (1 - \alpha_i)^2 + 
        \kappa \sum_{j\in \partial\{i\}} (1 - \alpha_j)^2
}
    ,
\end{align*}
}%
where $\mathcal {\tilde I}_i^c := \alpha_i \mathcal F_i^c$.
According to the above formula, 
a simple iteration algorithm can be obtained immediately, as summarized in Algorithm \ref{alg:naive}.
\begin{algorithm}
    \DontPrintSemicolon
    \SetAlgoNoLine
    \caption{Simple Foreground Refining Algorithm}
    \label{alg:naive}
    \SetKwFunction{Fun}{SimpleRefine}
    \SetKwProg{Fn}{Function}{:}{}
    \Fn{\Fun{$\mathcal{\tilde I}, \mathcal \alpha$, $\mathcal F_{\mathrm{init}}$, $\kappa$, $T$} } {
        $w \leftarrow$ width of $\mathcal F$
        ;
        $h \leftarrow$ height of $\mathcal F$
        \;
        $
        \hat{\mathcal F}(0) \leftarrow \mathcal F_{\mathrm{init}}$
        \;
        \For{$t \leftarrow 0$ \KwTo $T-1$}{
        \For{$i \leftarrow 1$ \KwTo $wh$}{
        \For{$c \leftarrow 1$ \KwTo $3$}{
    $
    \hat{\mathcal F}_i^c (t+1)
    \leftarrow
    \frac{
        \alpha_i
        \mathcal {\tilde I}_i^c
        +
        \kappa
         \sum\limits_{j \in \partial \{i\}} ((1-\alpha_i)^2 +(1 -\alpha_j)^2 ) \hat{\mathcal F}_j^c (t)
    }{
        \alpha_i^2 + 
        4\kappa
        (1 - \alpha_i)^2 + 
        \kappa
        \sum\limits_{j\in \partial\{i\}} (1 - \alpha_j)^2
}
    $
            \;
        }
    }
}
        \Return $\hat{\mathcal F} (T)$\;
    }
\end{algorithm}

In each iteration of Algorithm \ref{alg:naive}, the color in a pixel is only affected by its surrounding $4$ pixels.
Due to the slow speed of foreground color propagation,
Algorithm \ref{alg:naive} may 
have an extremely large mixing time.
Motivated by the fast multi-level foreground estimation method \cite{Germer2020FastMulti-Level},
we propose a multi-level refining procedure which first performs Algorithm \ref{alg:naive} on the downsampled image,
and gradually increases the image size to the original size.
We summarize the multi-level algorithm in Algorithm \ref{alg:multi-scale}.
In practice, the parameters in Algorithm \ref{alg:multi-scale} are $S= 6$, $\kappa = 1$, $T = 20$ and the resize operator is the bilinear interpolation.
\begin{algorithm}[t]
    \DontPrintSemicolon
    \SetAlgoNoLine
    \caption{Multi-Scale Foreground Refining Algorihtm}
    \label{alg:multi-scale}
    \KwIn{%
        Foreground $\mathcal F$; Opacity $\alpha$;
        Scale number $S$;
        Hyperparameter $\kappa$;
        Iteration number $T$;
    }
    \SetKwFunction{Fun}{SimpleRefine}
    \SetKwFunction{Resize}{Resize}
    $w \leftarrow$ width of $\mathcal F$
    ;
    $h \leftarrow$ height of $\mathcal F$
    \;
    $\mathcal F_{\mathrm{init}} \leftarrow \mathcal F$ 
    ;
    $\tilde {\mathcal I} \leftarrow \alpha \mathcal F$ 
    \;
\For{$s \leftarrow S - 1$ \KwTo $0$}{
    $\tilde {\mathcal I}_s \leftarrow $ \Resize{ $\tilde {\mathcal I}$ , $2^{-s}w$, $2^{-s}h$ }
    \;
    $\alpha_s \leftarrow $ \Resize{ $\alpha$ , $2^{-s}w$, $2^{-s}h$ }
    \;
    $\mathcal F_{\mathrm{init}} \leftarrow $ \Resize{ $\mathcal F_{\mathrm{init}}$ , $2^{-s}w$, $2^{-s}h$ }
    \;
    $\mathcal F_{\mathrm{init}} \leftarrow$ \Fun{ $\tilde {\mathcal I}_s$, ${ \alpha_s }$, $\mathcal F_{\mathrm{init}}$, $\kappa$, $2^s T$}  \;
}
\Return $\mathcal F_{\mathrm{init}}$\;
\end{algorithm}

In Fig. \ref{fig:four_and_four}, we illustrate the effect of re-estimation of $\mathcal F$ on an image in the training set of AIM \cite{Xu2017DeepImageMatting}.
It can be seen that, before re-estimation, the foreground has large variation when $\alpha_i$ is near $0$, which violates Assumption \ref{assumption1}.
After we re-estimate the foreground using the proposed algorithm, 
the foreground is much smoother when $\alpha_i$ is near $0$.
Also, the foreground should be continuous in the region $\{i: \alpha_i = 0\}$.


\subsection{Color Augmentation}
In this section, we propose a simple color augmentation method to augment the training data.
Given a foreground $\mathcal F$, We randomly generate an image $\tilde{\mathcal I}$ with a constant color $u$, i.e., $\tilde{\mathcal I}_i = u$ for all $i$, and 
a random number $w \in [0,1]$.
The augmentated foreground is defined as $\tilde{ \mathcal F} = w \mathcal F  + (1- w) \tilde{\mathcal I}$.
To see why this simple augmentation method can work, we note that the gradient of $\tilde{\mathcal F}$ is $\nabla \tilde{\mathcal F} = w \nabla \mathcal F$.
Since $w<1$, the augmented image $\tilde{\mathcal F}$ is smoother than $\mathcal F$.
Consequently, the augmented image is more consistent with Assumption \ref{assumption1}.

\subsection{Backpropagating Refinement}

\begin{figure*}[t]
  \centering
  \includegraphics[width = 1.90\columnwidth]{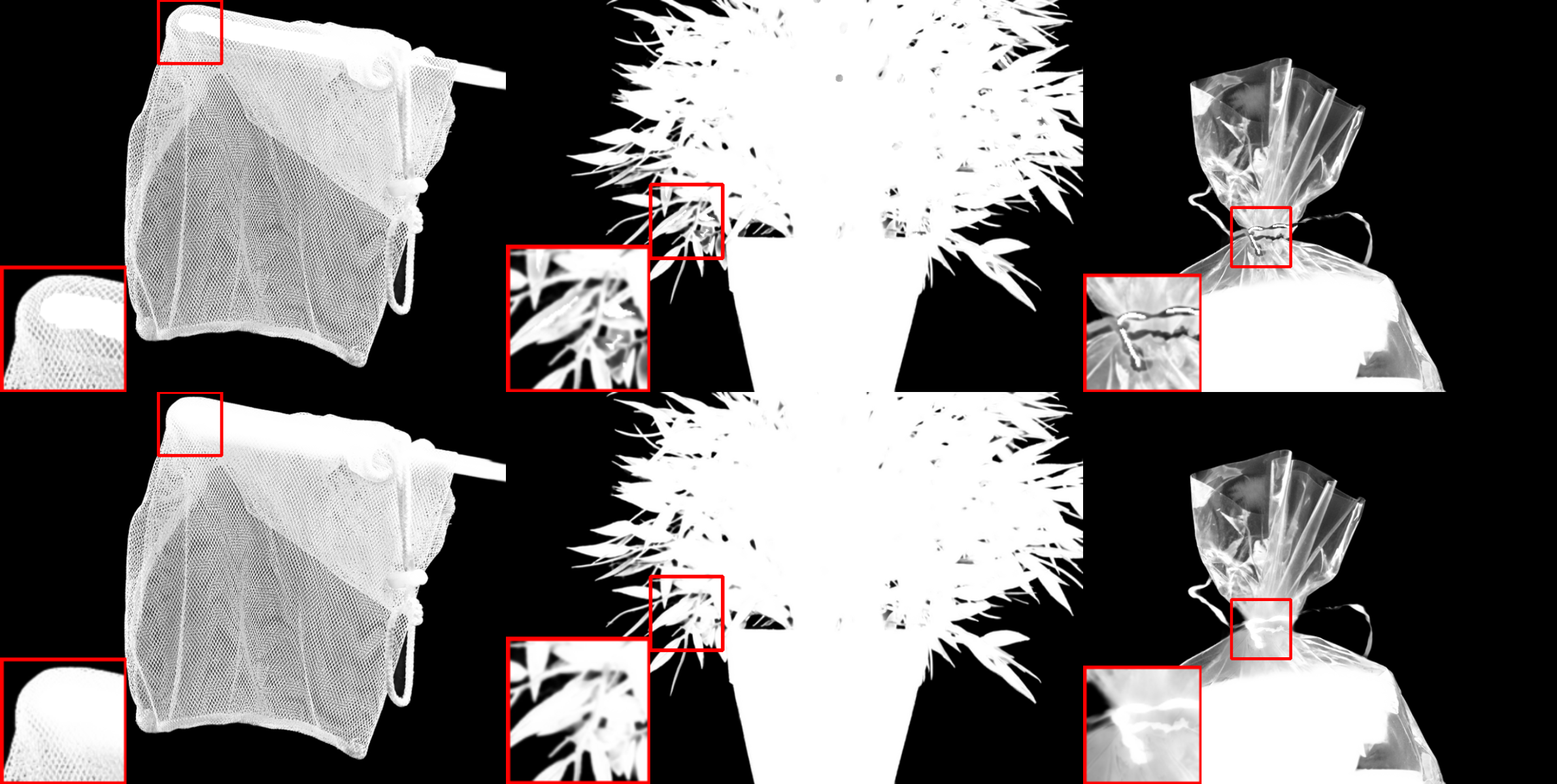}
  \caption{Illustration of the effect of BR.
      Figures in the first row are tested without BR.
      Figures in the second row are tested with BR.
  }
  \label{fig:666}
\end{figure*}

Now we consider the use of Assumption \ref{assumption2}
which
basically assumes that
the predicted opacity $\hat \alpha$ should take value $1$
on $\partial \{i: \mathcal T_i = 1\}$ and take value $0$ on $\partial \{i: \mathcal T_i = 0\}$. 
Intuitively, this assumption should be satisfied 
by any reasonable matting methods.
Surprisingly,
as illustrated in the first row of Fig. \ref{fig:666}, 
even for deep image matting models with goo performance,
the output of the network can violate Assumption \ref{assumption2}.
It shows that deep learning methods may not
automatically
guarantee that the known pixels annotated by the trimap has the correct result.
In fact,
similar phenomenon was previously observed in the field of interactive image segmentation \cite{Jang2019Interactive,Sofiiuk2020Rethinking}.

There are two possible causes of this phenomenon.
First, in the training phase of most existing methods, the trimaps are obtained by dilating the regions $\{i: \alpha_i > 0\}$ and $\{i: \alpha_i < 1\}$.
However, such trimaps can not fully mimick the input trimap generated by users.
Second, 
the information of sparse regions may not be fully extracted by convolutional neural networks.
Specifically, suppose $R$ is a connected component of the region $\{i: \mathcal T_i = 1\}$ with a very small area.
Then the important information provided by $R$ may be very likely to be ignored by convolutional neural networks.


Given a testing image $\mathcal I$ and its trimap $\mathcal T$,
if the output of the deep image matting model does not satisfy Assumption \ref{assumption2},
we would like to regulate the output to meet Assumption \ref{assumption2}.
To achieve this,
the idea is to use a backpropagating procedure to refine the result during the inference phase.
The use of backpropagating refinement in the inference phase
has achieved great success in the interactive segmentation task \cite{Jang2019Interactive,Sofiiuk2020Rethinking}.
To the best of our knowledge, 
the use of backpropagating refinement in deep image matting has never been explored before.

Our backpropagating refinement method works as follows.
For $2 \leq \ell \leq 5$,
the output of Decoder $\ell$ is processed by a Backpropagating Refinement Module (BRM)
whose
architecture is illustrated in Fig. \ref{fig:1}.
For BRM $\ell$, there are two trainable parameters $\BA$ and $\BB$ which are both tensors with dimension $(2^{8-\ell}, 2^{8-\ell}, C)$ where $C$ is the channel number of the input of BRM.
The output of BRM is $\exp( \mathscr I(\BA))\cdot \textrm{Input} + \mathscr I(\BB)$,
where  $\mathscr I$ is the interpolation operator such that $\mathscr I(\BA)$ and $\mathscr I(\BB)$ have the same dimension as the input.
In the training phase, the elements of $\BA$ and $\BB$ are fixed to $0$, and hence BRM is simply the identity operator.
In the inference phase,
we freeze all parameters other than the parameters in BRM.
Given a testing image,
we initialize the elements of $\BA$ and $\BB$ as $0$
and compute the output of the network $\hat \alpha_{\mathrm{init}}$.
After that, $\BA$ and $\BB$ are iteratively updated 
via gradient descent
to minimize the cost function
{\small
    \begin{align*}
        C_1(\hat \alpha, \mathcal T)
    +
    0.1
    C_2(\hat \alpha, \mathcal T)
,
\end{align*}
}%
where $\hat \alpha$ is the output of the network and 
{\small
    \begin{align*}
        C_1(\hat \alpha, \mathcal T)
        =&
        \frac{
            \sum_{i \in \partial\{i: \mathcal T_i = 0\} } \hat \alpha_i^2
            +
            \sum_{i \in \partial\{i: \mathcal T_i = 1\} } (1-\hat \alpha_i)^2
        }
        {
            \mathrm{Card}(\partial\{i: \mathcal T_i = 0\})
            +
            \mathrm{Card}(\partial\{i: \mathcal T_i = 1\})
        }
        ,
        \\
        C_2(\hat \alpha, \mathcal T)
        =&
        \left\{
        \frac{
            \sum_{i \in \{i: \mathcal T_i = 0.5 \}}|\hat \alpha_i - \hat \alpha_{\text{init},i} |
            }{
            \mathrm{Card}(\{i: \mathcal T_i = 0.5 \})
        }
        \right\}^2
        .
    \end{align*}
}%
The cost function $C_1$ regulates the output to meet Assumption \ref{assumption2}.
There may be only a small number of pixels which severely violate Assumption \ref{assumption2}.
It is known that the $L_2$ loss is sensible to outliers.
Hence in $C_1$, we use the $L_2$ loss which pay more attention to the pixels that largely violate Assumption \ref{assumption2}.
On the other hand, the cost function $C_2$ makes sure the output does not largely deviate from the original output of the network.
It is known that the $L_1$ loss is robust against a sparse set of outliers.
Hence in $C_2$, 
the squared $L_1$ loss ensures that the majority of pixels are not far away from the original output.

Note that the BRMs are all in the decoder.
Hence we only need to forward the full network once and then the backward and forward computation can be restricted to the decoder.
This reduces the computational cost.
In practice, we set the iteration number to be $ 100$ with learning rate $ 20$.


\begin{table}[b]
\centering
    \caption{
        Ablation studies on Composition-1k testing dataset.
        IH: input height $h$ of training image patches.
        CF: closed-form foreground estimation method in \cite{Levin2008AClosed-Form}.
}
    \label{table1}
\setlength\tabcolsep{4pt}
\begin{tabular}{|cccc|cccc|}
    \hline
    IH& RF & CA & TTA
      & SAD & MSE & Grad & Conn
  \\
  \hline
    320 & &  & 
            & 32.4 & 0.0074 & 11.6 & 29.6
    \\
    480 & &  &
            & 30.0 & 0.0067 & 10.3 & 26.6
    \\
    640 & &  &
            & 29.0 & 0.0062 & 10.3 & 25.4
    \\
    640 & CF & &
            & 28.9 & 0.0058 & 10.2 & 25.4
    \\
    640 & \checkmark &  & 
            & 27.6 & 0.0057 & 10.1 & 23.6
    \\
    640 & & \checkmark & 
            & 27.9 & 0.0056 & 9.94 & 23.8
    \\
    640 & \checkmark & \checkmark & 
            & 25.9 & 0.0054 & 9.25 & 21.5
            \\
    640 & \checkmark & \checkmark &  \checkmark
        & \textbf{24.6} & \textbf{0.0045} & \textbf{8.13} & \textbf{19.9}
            \\
\hline
\end{tabular}
\end{table}

\begin{table}[t]
\centering
    \caption{
        Quantitative results on Composition-1k.
}
    \label{table2}
\setlength\tabcolsep{4pt}
\begin{tabular}{|l|cccc|}
    \hline
    Methods & SAD & MSE & Grad & Conn  
  \\
    \hline
    Closed-form \cite{Levin2008AClosed-Form} & 168.1 & 0.091 &126.9 & 167.9
    \\
    DIM \cite{Xu2017DeepImageMatting} & 50.4 & 0.014 & 31.0 & 50.8
    \\
    IndexNet \cite{Lu2019IndicesMatter} & 45.8 & 0.013 & 25.9 & 43.7
    \\
    SampleNet \cite{Tang2019Learning-Based} & 40.4 & 0.0099 & - & -
    \\
    Context-aware \cite{Houjj2019ContextAwareImageMatting} & 35.8 & 0.0082 & 17.3 & 33.2
    \\
    GCA \cite{Li2020NaturalImageMatting} 
    & 35.3 & 0.0091 & 16.9 & 32.5
    \\
    HDMatt \cite{Yu2021High-ResolutionDeepImageMatting}
    & 33.5 & 0.0073 & 14.5 & 29.9
    \\
    A$^2$U \cite{Dai2021LearningAffinityAware}
    & 32.2 & 0.0082 & 16.4 & 29.3
    \\
    TIMI-Net \cite{Liu2021TripartiteInformationMining} & 29.1 & 0.0060 & 11.5 & 25.4
    \\
    SIM \cite{Sun2021SemanticImageMatting} & 28.0 & 0.0058 & 10.8 & 24.8
    \\
    FBA \cite{Forte2020FBA} & 26.4 & 0.0054 & 10.6 & 21.5
    \\
    FBA + TTA \cite{Forte2020FBA} &  25.8 & 0.0052 & 10.6 & 20.8
    \\
    LFPNet \cite{Liu2021Long-RangeFeaturePropagating} 
                                  & 23.6 & 0.0041 & 8.4 & 18.5
                                  \\
    LFPNet + TTA \cite{Liu2021Long-RangeFeaturePropagating} 
                                  & \textbf{22.4} & \textbf{0.0036} & \textbf{7.6} & \textbf{17.1}
    \\
    \hline
    RF + CA (Ours) &
            25.9 & 0.0054 & 9.25 & 21.5
            \\
    RF + CA + TTA (Ours) &
            24.6 & 0.0045 & 8.13 & 19.9
    \\
    \hline
\end{tabular}
\end{table}

\begin{figure*}[t]
  \centering
  \includegraphics[width = 1.90\columnwidth]{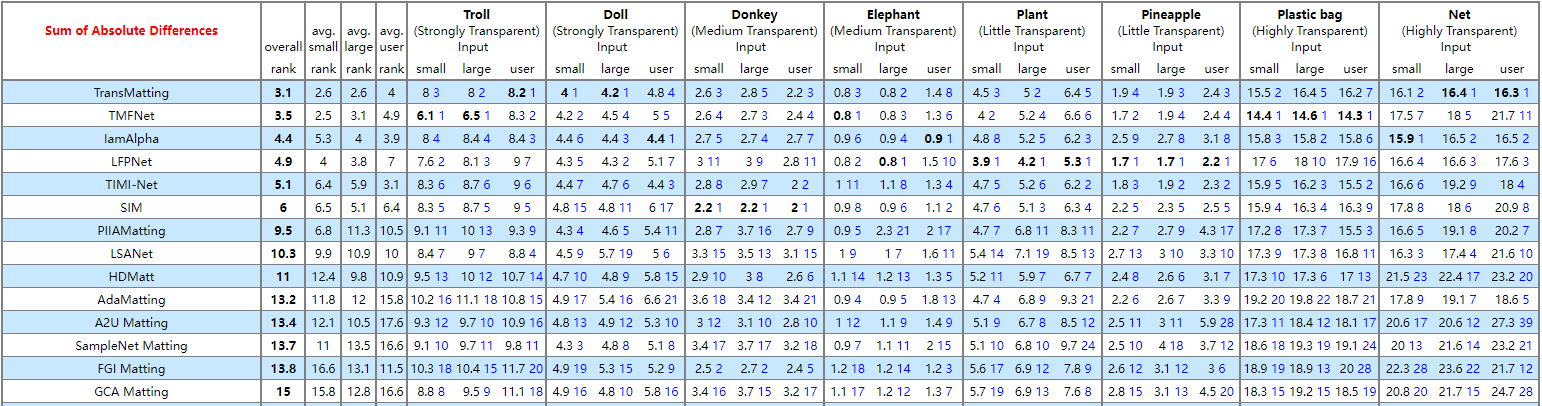}
  \caption{Performance of the proposed method with BR on the AlaphaMatting testing set.}
  \label{fig:100}
\end{figure*}

\begin{figure*}[t]
  \centering
  \includegraphics[width = 1.90\columnwidth]{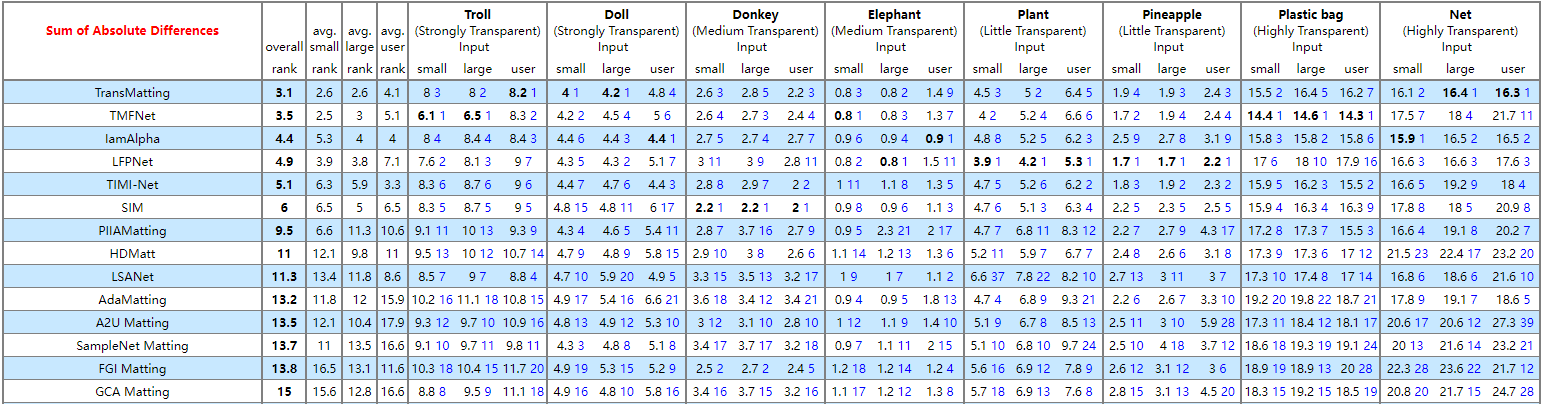}
  \caption{Performance of the proposed method without BR on the AlaphaMatting testing set.}
  \label{fig:101}
\end{figure*}

\section{Experiments}
In this section, we report experimental results of the proposed method.
The source code is publicly available at 
\texttt{https://github.com/kfeng123/LSA-Matting}.

\textbf{Training data.}
For all experiments,
the encoder of the network is pretrained on ImageNet \cite{Deng2009ImageNet},
and we train the network on the AIM training set \cite{Xu2017DeepImageMatting} which consists of $431$ distinct foreground images companioned with their alpha matte and
$43,100$ background images from MS-COCO dataset \cite{Lin2014MS-COCO}.
No additional data is used.
Instead of compositing each foreground image with a prespecified background image, we composite each foreground image with a randomly selected background in each iteration of training phase.

\textbf{Data augmentation.}
Our data augmentation procedure for the baseline model is as follows.
First, with probability $0.5$, we randomly rotate the pair $(\mathcal F, \alpha)$ and $\mathcal B$.
Then we randomly crop $(\mathcal F, \alpha)$ and $\mathcal B$, centered at a random pixel in the region $\{i: \alpha > 0\}$ with size $k \times k$ where $k$ is randomly chosen in $[\frac{3}{4} h, \frac{5}{4} h]$, and then resize it to $h \times h$ where $h$ is the height of the input image patches in training phase.
After that, with probability $0.5$, we flip $(\mathcal F, \alpha)$ and $\mathcal B$.
With probability $0.3$, we transform $\alpha$ to
$\alpha^{\gamma}$ or $1- (1- \alpha)^\gamma$ with equal probablity where $\gamma$ is randomly chosen in $[0.5, 1.5]$.
With probability $0.3$, we transform $F$ to $1-F$.
With probability $0.3$, we randomly permute the $3$ channels of $F$.
Finally, we generate the trimap by dilating the regions $\{i:\alpha_i > 0\}$ and $\{i: \alpha <1\}$ by random numbers from $1$ to $24$.


\textbf{Implementation details.}
We use Adam optimizer \cite{Kingma2015Adam}
with initial learning rate $5\cdot 10^{-5}$.
We halve the learning rate every $20$ epochs.
The model is trained for $100$ epochs with
batch size $16$
and weight decay $10^{-4}$.

\textbf{Evaluation Metrics.}
Following \cite{Xu2017DeepImageMatting,Houjj2019ContextAwareImageMatting}, we use the following metrics to evaluate matting methods:
the sum of absolute differences (SAD), the mean square errors (MSE), the gradient errors (Grad) and the connectivity errors (Conn).


\subsection{Results on Composition-1k Testing Dataset}
In this section, we report evaluation results on the Composition-1k testing dataset \cite{Xu2017DeepImageMatting}
to illustrate the effectiveness
of two of the proposed techniques: re-estimated foregrounds (RF) and color augmentation (CA).
We will evaluate the third technique in the next subsection since this technique is time-consuming in this dataset.
In addition,
we also test our models with Test Time Augmentation (TTA).
TTA was previously used by \cite{Tang2019Learning-Based,Forte2020FBA} to improve the performance of the network.
When TTA is used, the image is tested at three scales $0.8$, $1$, $1.25$.
For each scale, the image is rotated by $k\pi / 2$, $k  =0, 1, 2, 3$, and is flipped to generate $8$ images.
We run the model on these $3 \times 8 = 24$ images and use the averaged output as the final result.


\textbf{Ablation results.}
Table \ref{table1} lists the ablation results of the proposed techniques.
The results indicate the following phenomenons.
First,
while using the closed-form foreground estimation method in \cite{Levin2008AClosed-Form}
can improve the performance,
the proposed RF can lead to even better results.
Second,
the proposed CA can also lead to great improvement.
Also,
our results give quantitative characterization of the known facts
that
large input size in the training phase 
and TTA in the inference phase can lead to 
significant improvement of the performance.


\textbf{Comparisons with the state-of-the-art methods.}
In table \ref{table2}, we list the performances of some recent methods on Composition-1k.
It can be seen that 
the methods of \cite{Liu2021Long-RangeFeaturePropagating}
are the only ones whose overall performance is better than ours.
However, 
\cite{Liu2021Long-RangeFeaturePropagating} used additional data to pretrain their network.
In comparison, no extra data is used in our model.
Thus, our model achieves new state-of-the-art performance for all $4$ metrics among all methods that are solely trained on the AIM training set.



\subsection{Results on AlphaMatting Dataset}

\begin{table}[b]
\centering
    \caption{
        Ablation studies on AlphaMatting training set.
}
    \label{tablewuwuwu}
\setlength\tabcolsep{4pt}
\begin{tabular}{|l|cccc|}
    \hline
    Methods & SAD & MSE & Grad & Conn  
  \\
    \hline
    Without BR
            & 2.90 &0.00540 & 2.47 & 2.45
    \\
    With BR
            & 2.87 & 0.00532 & 2.43 & 2.42
    \\
    \hline
\end{tabular}
\end{table}

While our model is trained
on the AIM training set, we also test our model on AlphaMatting dataset \cite{Rhemann2009APerceptually}.
The data distribution of AlphaMatting dataset is different from AIM dataset.
To increase the generalization ability of our model,
we add an additional data augmentation technique in the training phase, that is, with probability $0.3$, the foreground and background are blurred.
Also, we train the model for $200$ epochs.
For all results, techniques RF, CA and TTA are used by default.

\textbf{Ablation results.}
We evaluate the proposed backpropagating refinement (BR) technique on AlphaMatting training set.
The results, listed in Table \ref{tablewuwuwu},
show that BR can improve all $4$ metrics.
This verifies the effectiveness of BR.

\textbf{Comparisons with the state-of-the-art methods.}
We evaluate the proposed method on AlphaMatting testing set.
The performances of the proposed method (denoted as LSANet)
and some competing methods are illustrated in 
Fig. \ref{fig:100} and
Fig. \ref{fig:101}.
It can be seen that BR leads to an improvement of the ranking.
While the proposed method does not rank among the top methods, it outperforms some recent methods, including \cite{Tang2019Learning-Based,Li2020NaturalImageMatting,Dai2021LearningAffinityAware,Yu2021High-ResolutionDeepImageMatting}.
Also, the proposed method uses a simple architecture and is trained solely on AIM dataset.
Overall, the performance of the proposed method is promising.


\section{Conclusion}
In this paper, 
    we investigated the use of local smoothness assumptions in deep image matting and proposed three techniques which can improve the performance of the deep image matting model significantly.
    We adopted a simple network and trained the model on AIM training data.
    Extensive experiments verified the effectiveness of our method.



\bibliographystyle{IEEEbib}
\bibliography{mybibfile}

\end{document}